%% file: acl2023.tex
\pdfoutput=1

\documentclass[11pt]{article}

\usepackage{ACL2023}

\usepackage{times}
\usepackage{latexsym}

\usepackage[T1]{fontenc}

\usepackage[utf8]{inputenc}

\usepackage{microtype}

\usepackage{inconsolata}
\usepackage{subcaption}
\usepackage{booktabs}
\usepackage{graphicx}
\usepackage{dingbat}
\usepackage{multirow}
\usepackage{tikz}

\title{How do humans perceive adversarial text? A reality check on the validity and naturalness of word-based adversarial attacks}

\author{Salijona Dyrmishi \\
  University of Luxembourg\\
  \texttt{salijona.dyrmishi@uni.lu} \\
  \And
 Salah Ghamizi \\
 University of Luxembourg  \\
  \texttt{salah.ghamizi@uni.lu} \\
  \And
Maxime Cordy \\
 University of Luxembourg  \\
  \texttt{maxime.cordy@uni.lu} \\
  }

\begin{document}
\maketitle
\begin{abstract}

Natural Language Processing (NLP) models based on Machine Learning (ML) are susceptible to adversarial attacks -- malicious algorithms that imperceptibly modify input text to force models into making incorrect predictions. However, evaluations of these attacks ignore the property of imperceptibility or study it under limited settings. This entails that adversarial perturbations would not pass any human quality gate and do not represent real threats to human-checked NLP systems. To bypass this limitation and enable proper assessment (and later, improvement) of NLP model robustness, we have surveyed 378 human participants about the perceptibility of text adversarial examples produced by state-of-the-art methods. Our results underline that existing text attacks are impractical in real-world scenarios where humans are involved. This contrasts with previous smaller-scale human studies, which reported overly optimistic conclusions regarding attack success. Through our work, we hope to position human perceptibility as a first-class success criterion for text attacks, and provide guidance for research to build effective attack algorithms and, in turn, design appropriate defence mechanisms.

\end{abstract}


\input{Introduction.tex}

\input{Motivation.tex}
\input{RQs.tex}

\input{Design.tex}
\input{results_RQ1.tex}
\input{results_RQ2.tex}

\input{results_RQ3.tex}
\input{results_misc.tex}

\input{Discussion.tex}
\input{Conclusion.tex}

\input{Limitations.tex}

\section*{Acknowledgements}
Salijona Dyrmishi's work is supported by the Luxembourg National Research Funds (FNR) AFR Grant 14585105.

\bibliography{references}
\bibliographystyle{acl_natbib}

\input{Appendix}

\end{document}

%% file: Introduction.tex
\section{Introduction}

Like many other machine learning models, Natural Language Processing (NLP) models are susceptible to adversarial attacks. In NLP, these attacks aim to cause failures (e.g. incorrect decisions) in the model by slightly perturbing the input text in such a way that its original meaning is preserved. 

Research has reported on the potential of adversarial attacks to affect real-world models interacting with human users, such as Google's Perspective and Facebook's fastText \cite{Li2018TextBuggerGA}) More generally, these attacks cover various learning tasks including classification and seq2seq  (fake news \cite{li-etal-2020-bert-attack},  toxic content \cite{Li2018TextBuggerGA}, spam messages \cite{kuchipudi2020adversarial}), style transfer \cite{qi-etal-2021-mind} and machine translation \cite{michel-etal-2019-evaluation}). 


It is critical to properly assess model robustness against adversarial attacks to design relevant defence mechanisms.
This is why research has investigated different attack algorithms based on paraphrasing \cite{iyyer2018adversarial}, character-level \cite{gao2018black, pruthi-etal-2019-combating} and word-level \cite{garg-ramakrishnan-2020-bae,ren2019generating} perturbations, and made these algorithms available in standardized libraries \cite{morris2020textattack,zeng2020openattack}.

For the many NLP systems that interact with humans, we argue that \emph{effective adversarial attacks should produce text that is both \textbf{valid} and \textbf{natural}}. Validity refers to the property that humans perceive the same semantic properties of interest\footnote{In the case of classification tasks, these semantics properties boil down to the class labels.} for an adversarial text as for the original text it was produced from. Naturalness refers to the perception that an adversarial text was produced by humans. Adversarial texts that are invalid and/or unnatural can still cause failed NLP model decisions, however, their ultimate effect on humans is negligible because they would fail to convey the intended meaning (e.g. hate speech that is not perceived as hateful) or they would be suspected to be computer-generated (e.g., a phishing email using awkward vocabulary and grammar).


Unfortunately, the scientific literature on adversarial text attacks has neglected (and sometimes ignored) the inclusion of human perception as an essential evaluation criterion -- see Table \ref{tab:introduction}. We found that (i) 3  studies do not include humans at all in their evaluation; (ii) merely 12 studies consider naturalness, and they only do so under limited settings. Indeed, these studies involve a single attack, one or two naturalness criteria, less than 10 participants, and they disregard the impact of parameters and factors like perturbation size and language proficiency. Instead, the studies rely on automated metrics (i.e cosine distance to measure semantic similarity), but these are not suitable proxies for human perception \cite{morris-etal-2020-reevaluating}.

The absence of systematic analysis of adversarial texts \emph{as perceived by humans} risks leading to overestimation of their semantic quality and, in turn, to fallacious model robustness assessment and misguidance during the design of defences. This was hinted in the seminal work from \citet{morris-etal-2020-reevaluating}, where a 10-participant survey on one dataset and two attacks revealed a discrepancy between the human-perceived naturalness of adversarial examples.


Therefore, in this paper, we present the first extensive study that evaluates the human-perceived validity and naturalness of adversarial texts. We surveyed 378 participants in assessing, based on five criteria, over 3000 texts (original and adversarial) coming from three datasets and produced by nine state-of-the-art attacks. 

Our investigations first reveal that the participants would classify 28.14\% of adversarial examples into a different class than the original example. This means that the adversarial perturbations change human understanding of the modified text and, thus, fail to achieve their purpose. Irrespective of the classification task, participants detect 60.3\% of adversarial examples as computer-altered; they can even identify 52.38\% of the exact altered word. These findings contrast the overly optimistic conclusions regarding attack success rates from previous small-scale human studies. Our results underline that existing attacks are not effective in real-world scenarios where humans interact with NLP systems. Through our work, we hope to position human perception as a first-class success criterion for text attacks, and provide guidance for research to build effective attack algorithms and, in turn, design appropriate defence mechanisms.

\begin{table*}[t]
\small
\centering
\begin{tabular}{@{}l|c|ccccc|c|c@{}}
\toprule
\multicolumn{1}{c|}{Attack name/paper}               & 
\multicolumn{1}{c|}{Type}        & 
\multicolumn{5}{c|}{Evaluation}                                                                                                                             & Participants         & \begin{tabular}[c]{@{}c@{}}Attacks \\ studied\end{tabular} \\ \midrule
\multicolumn{1}{c|}{}                     & \multicolumn{1}{l|}{}            & \multicolumn{1}{c|}{Validity}   & \multicolumn{4}{c|}{Naturalness}                                                                                          &                       &                                                            \\
                                          & \multicolumn{1}{l|}{}            & \multicolumn{1}{l|}{}           & S.                              & D.                     & G.                             & M.                              & \multicolumn{1}{l|}{} & \multicolumn{1}{l}{}                                       \\ \cmidrule(lr){3-7}
Hotflip \cite{ebrahimi-etal-2018-hotflip}        & \multirow{12}{*}{Word based}     & \multicolumn{1}{c|}{\checkmark} & X                              & X                     & X                              & X                               & 3                     & 1                                                          \\
Alzantot\cite{alzantot-etal-2018-generating}     &                                  & \multicolumn{1}{c|}{\checkmark} & X                              & X                     & X                              & X                               & 20                    & 1                                                          \\
Input-reduction\cite{feng-etal-2018-pathologies} &                                  & \multicolumn{1}{c|}{\checkmark} & X                              & X                     & X                              & X                               & N/A                   & 1                                                          \\
Kuleshov\cite{kuleshov2018adversarial}    &                                  & \multicolumn{1}{c|}{\checkmark} & X                              & X                     & X                              & X                               & 5                     & 1                                                          \\
Bae\cite{garg-ramakrishnan-2020-bae}                     &                                  & \multicolumn{1}{c|}{\checkmark} & \checkmark                     & X                     & \checkmark                     & X                               & 3                     & 2                                                          \\
Pwws\cite{ren2019generating}              &                                  & \multicolumn{1}{c|}{\checkmark} & \checkmark                     & X                     & X                              & X                               & 6                     & 1                                                          \\
Textfooler \cite{jin2020bert}             &                                  & \multicolumn{1}{c|}{\checkmark} & X                              & X                     & \checkmark                     & \checkmark                      & 2                     & 1                                                          \\
Bert-attack\cite{li-etal-2020-bert-attack}              &                                  & \multicolumn{1}{c|}{\checkmark} & X                              & X                     & \checkmark                     & X                               & 3                     & 1                                                          \\
Clare \cite{li-etal-2021-contextualized}         &                                  & \multicolumn{1}{c|}{\checkmark} & X                              & X                     & X                              & X                               & 5                     & 2                                                          \\
PSO \cite{zang2020word}                   &                                  & \multicolumn{1}{c|}{\checkmark} & \checkmark                     & X                     & X                              & X                               & 3                     & 1                                                          \\
Fast-alzantot \cite{jia-etal-2019-certified}     &                                  & \multicolumn{1}{c|}{X}          & X                              & X                     & X                              & X                               & 0                     & 0                                                          \\
IGA \cite{wang2019natural}                &                                  & \multicolumn{1}{c|}{X}          & X                              & X                     & X                              & X                               & 0                     & 0                                                          \\ \midrule
Textbugger \cite{Li2018TextBuggerGA}         & \multirow{3}{*}{Character based} & \multicolumn{1}{c|}{\checkmark} & X                              & \checkmark            & X                              & X                               & 297                   & 1                                                          \\
Pruthi \cite{pruthi-etal-2019-combating}          &                                  & \multicolumn{1}{c|}{\checkmark} & X                              & X                     & X                              & X                               & N/A                  & 1                                                          \\
DeepWordBug \cite{gao2018black}           &                                  & \multicolumn{1}{c|}{X}          & X                              & X                     & X                              & X                               & 0                     & 0                                                          \\ \midrule
\citet{morris-etal-2020-reevaluating}                             & \multirow{2}{*}{Independent}            & \multicolumn{1}{c|}{X}          & \checkmark & X & \checkmark & \checkmark & 10                    & 2                                                          \\
\textbf{Our study}        &                                    & \multicolumn{1}{c|}{\checkmark} & \checkmark                     & \checkmark            & \checkmark                     & \checkmark                      & 378                   & 9                                                          \\ \bottomrule
\end{tabular}
\caption{Human evaluation performed on quality of adversarial examples by existing literature. The terms abbreviated are Suspiciousness(S.), Detectability(D.), Grammaticality(G.), Meaning(M.).   N/A indicates information is not available.}
\label{tab:introduction}
\end{table*}

%% file: Motivation.tex
\begin{figure}[ht!]
\includegraphics[width=\linewidth]{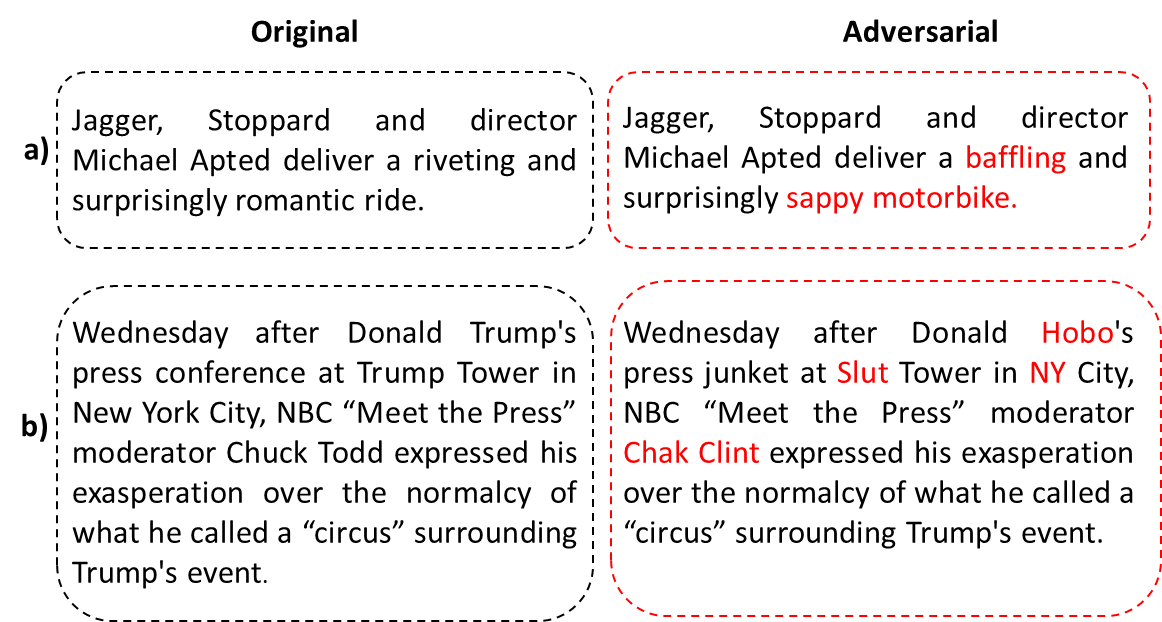}
\caption{Adversarial examples against NLP model, with perturbations in red. a) Invalid adversarial example generated by~\cite{morris-etal-2020-reevaluating}. b) Unnatural adversarial example generated by ~\citet{ali2021all}.}
\label{fig:adv_example_fake_news}
\end{figure}

\section{Motivation}

Consider the example of fake news shown in Figure~\ref{fig:adv_example_fake_news}b. (``\emph{Original}''). \citet{ali2021all} have shown that this example is detected by existing fake news detectors based on NLP machine learning models. However, the same authors have also revealed that, if one changes specific words to produce a new sentence (\emph{``Adversarial''}), the same detector would fail to recognize the modified sentence as fake news. This means that fake news could ultimately reach human eyes and propagate. 



Fortunately, fake news -- like hate speech, spam, phishing, and many other malicious text contents -- ultimately targets human eyes and has not only to bypass automated quality gates (such as detectors) but also fool human understanding and judgment. Indeed, to achieve their goal of propagating erroneous information, adversarial fake news should still relay wrong information -- they should be ``valid'' fake news -- and be perceived as a text seemingly written by humans -- i.e. they should be ``natural''. The fake news example from Figure~\ref{fig:adv_example_fake_news} is unnatural because it uses irrelevant proper nouns like "Slut Tower" or "Donald Hobo" that do not exist in reality, and this makes the fake news ineffective. We, therefore, argue that invalid and/or unnatural examples do not constitute relevant threats.

Thus, the goal of adversarial text attacks becomes to produce examples that change model decision and are
perceived by humans as valid and natural. 
Our study aims to assess, using human evaluators, whether state-of-the-art text adversarial attacks meet this goal. The answer to this question remains unknown today because, as revealed by our survey of existing attacks (see Table \ref{tab:introduction}), only six papers cover both validity and naturalness, five of them do so with less than 10 human participants, and Textbugger \cite{Li2018TextBuggerGA} that has the largest number of participants assesses naturalness only at word level, not sentence level. 
\textit{Nevertheless, all these papers evaluate the effectiveness of the specific attack they introduce (rarely with another baseline) and there is a lack of standardized studies considering them all.}

For our study, the validity and naturalness requirements led us to consider word-based attacks. Indeed, character-based attacks are easily detectable by humans and are even reversible with spelling and grammar check methods \cite{sakaguchi-etal-2017-grammatical}. In word-based attacks, the size of the perturbation $\delta$ is typically defined as the number of modified words.

%% file: RQs.tex
\section{Research questions and metrics}
\subsection{Research questions}
\label{subsec:RQs}
 
 
Our study firstly investigates the validity of adversarial examples as perceived by humans.

\begin{description}
\item[\bf RQ1 (Validity):] \em Are adversarial examples valid according to human perception?
\end{description}

Validity is the ability of the adversarial example to preserve the class label given to the original text \cite{chen-etal-2022-adversarial}. Figure \ref{fig:adv_example_fake_news}a) illustrates a case of an invalid adversarial example, which changes the positive sentiment of the original example. Thus, we aim to compare the label that human participants would give to an adversarial example with the label of the original example. To determine the original label, we use as a reference the ``ground truth'' label indicated in the original datasets used in our experiments -- that is, we assume that this original label is the most likely to be given by human evaluators. To validate this assumption, our study also confronts participants to original examples and checks if they correctly classify these examples (Section \ref{sec:validity}). A statistical difference between humans' accuracy on adversarial examples compared to original examples would indicate that a significant portion of adversarial examples is invalid.

In addition to validity, we study next the degree to which adversarial texts are natural. 
\begin{description}
\item[\bf RQ2 (Naturalness):] \em  Are adversarial examples natural?
\end{description}

To answer this question, we measure the ability of humans to suspect that a piece of text has been computer altered (with adversarial perturbations). An adversarial example is thus evaluated as less natural, the more it raises \emph{suspicion} (to have been altered) among the participants.

The suspicion that a text seems computer-altered might arise from different sources, for example the use of specific words, typos, lack of semantic coherence etc. Thus, in addition to evaluating \emph{suspiciousness}, we refine our analysis in order to unveil some reasons why humans may found an adversarial text to be suspicious. We investigate three additional naturalness criteria:
\begin{itemize}
    \item \emph{Detectability} is the degree to which humans can recognize which words of a given adversarial sentence we altered. High detectability would indicate that the choice of words significantly affect the naturalness of these examples (or lack thereof). We assess detectability in two settings: wherein humans do not know how many words have been altered (unknown |$\delta$|)) and wherein they know the exact number of altered words (known |$\delta$|).
    
    \item \emph{Grammaticality} is the degree to which an adversarial text respects the rules of grammar. The presence of grammar errors in a text might raise the suspicion of human evaluators. However, grammar errors may also occur in original (human-written) text. 
    Therefore, we study both the total number of grammar errors in adversarial examples (``error presence''), and the number of introduced errors compared to original texts (``error introduction''). The latter is a better evaluator for the quality of generated adversarial text.  A high relative amount of grammar errors could explain the suspiciousness of the adversarial examples (or lack thereof). 

    \item \emph{Meaningfulness} is the degree to which the adversarial text clearly communicates a message that is understandable by the reader. We assess the meaningfulness of adversarial text first in isolation (``clarity'')), and then check whether humans believe the meaning of the original text has been preserved under the adversarial perturbation (``preservation''). We hypothesize that adversarial texts with significantly altered meanings are more suspicious. 
\end{itemize}


Finally, because the perturbation size is known to impact success rate and human perceptibility of adversarial attacks in other domains \cite{constrained_adversarials,dyrmishi2022empirical}, we investigate the relationship between the number of altered words and validity/naturalness.

\begin{description}
\item[\bf RQ3:] \em  How does perturbation size impact the validity and naturalness of adversarial examples?
\end{description}

Although there is a general acceptance that lower perturbation sizes are preferred, the actual magnitude of the effect that perturbation size causes on text perception has not been studied before.  

\subsection{Reported metrics}

Throughout our study, we compute different metrics for each attack separately and all attacks altogether.

\textbf{Validity:} the percentage of human-assigned labels to adversarial text that match the ground truth provided with the datasets. 

\textbf{Suspiciousness:} the percentage of adversarial texts recognized as ``computer altered".

\textbf{Detectability:} the percentage of perturbed words in an adversarial text that are detected as modified.

\textbf{Grammaticality:} the percentage of adversarial texts where human evaluators detected present errors (errors introduced by the attack), did not detect or were not sure. 

\textbf{Meaningfulness:}  the average value of clarity of meaning and meaning preservation, as measured on a 1-4 Likert scale (the Likert scale options are given in Figure \ref{fig:questionnaire}).

\subsection{Statistical tests}
\label{subsec:test}

To assess the significance of differences we observe, we rely on different statistical tests chosen based on the concerned metrics.
\begin{itemize}
    \item \textit{Proportion tests} are used for validity and suspicion, because they are measured as proportions.
    \item \textit{Mann Whitney U tests} are used for detectability, grammaticality and meaningfulness because their data are ordinal and may not follow a normal distribution (which this test does not assume). We compute the standardized Z value because our data samples are larger than 30, and the test statistic $U$ is roughly normally distributed.
    \item \textit{Pearson correlation tests} are used to assess the existence of linear correlations between the perturbation size and validity/naturalness.  
\end{itemize}
We perform all these tests with a significance level of $\alpha=0.01$.

%% file: Design.tex
\section{Study design}

\subsection{Adversarial texts}
\label{dataset}
To generate the adversarial texts presented to participants, we used the TextAttack library \cite{morris2020textattack}, which is regularly kept up to date with state-of-the-art attacks, including word-based ones.

\subsubsection{Attacks}
In total, we used nine word-based attacks from the library. Three of them(  \textit{BERTAttack} \cite{li-etal-2020-bert-attack}, \textit{BAE}\cite{garg-ramakrishnan-2020-bae}, \textit{CLARE}\cite{li-etal-2021-contextualized}) belong to the family of attacks that uses masked language models to introduce perturbations to the original text. Three others (\textit{FGA}\cite{jia-etal-2019-certified},  \textit{IGA}\cite{wang2019natural}, \textit{PSO}\cite{zang2020word}) use evolutionary algorithms to evolve the original text towards an adversarial one.  The remaining three (\textit{Kuleshov}\cite{kuleshov2018adversarial}, \textit{PWWS}\cite{ren2019generating}, \textit{TextFooler}\cite{jin2020bert}) use greedy search strategies. For all the attacks, we used the default parameters provided by the original authors. We excluded only Hotflip  attack because it was not compatible with the latest Bert-based models and Alzantot attack, for which we used its improved and faster version \textit{FGA}. You can refer to Table \ref{tab:introduction} for details related to the human study performed by the original authors. 

\subsection{Datasets}
We attacked models trained on three sentiment analysis datasets:  IMDB movie reviews \cite{maas-EtAl:2011:ACL-HLT2011}, Rotten Tomatoes movie reviews \cite{Pang+Lee:05a} and Yelp polarity service reviews \cite{zhang2015character}. We reuse the already available DistilBERT models in the TextAttack library that are trained on these three datasets. Sentiment analysis is a relevant task to assess validity and naturalness, and is easily understandable by any participant, even without domain knowledge. We limited the study to only one task to avoid the extra burden of switching between tasks for the participants. We include this choice in the section Limitations as a study with diverse tasks and datasets would be interesting (i.e datasets with more formal language). 

On each dataset, we ran the selected nine word-level attacks, which resulted in 25 283 successful adversarial examples in total. 





\subsection{Questionnaire}
\label{sec:survey}

\begin{figure}[ht!]
\includegraphics[width=\linewidth]{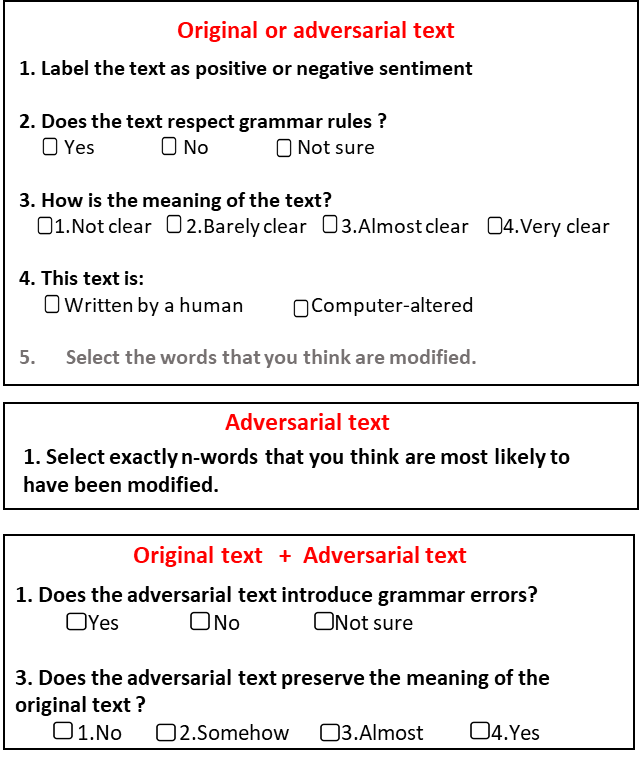}
\caption{The online questionnaire structure.}
\label{fig:questionnaire}
\end{figure}

We collected the data using an online questionnaire with three parts, presented in Figure \ref{fig:questionnaire}. The beginning of the questionnaire  contains the description of computer-altered text as "\textit{a text altered automatically by a program by replacing some words with others}".  We do not use the term ``adversarial examples'' to make the questionnaire accessible to non-technical audiences and avoid biases. We do not provide any hints to participants about the word replacement strategy (i.e. synonym replacement). In addition to this explanation, we  clarify to the participants the intended use of the data collected from this study.

The first part of the questionnaire shows examples in isolation and without extra information. It contains questions about validity, suspiciousness, detectability (unlimited choices), grammaticality (presence of grammar errors), and meaningfulness (clarity). We display only one text at a time, and each participant receives five random adversarial texts shuffled with five random original texts. We exclude the five original texts used as the initial point for the adversarial generation process, to ensure that participants do not look at two versions of the same text. Question number 5 on detectability will appear only if the participant answers "computer altered" to question 4. 


The second part focuses on detectability (exact number). Adversarial examples and their exact number $n$ of perturbed words are shown, and participants have to choose the $n$ words they believe have been altered. Each participant evaluates four adversarial examples they did not see in the first questionnaire part. 

The third part shows original and adversarial examples together. It contains questions about grammaticality (errors introduction) and meaning (preservation).  Each participant sees the same four adversarial examples (s)he had in the second part and their corresponding original examples. 

For each participant, we have (randomly) selected the displayed adversarial examples in order to ensure a balance between the different attacks and perturbation sizes. Each participant sees nine adversarial examples in total (one per attack) with different perturbation sizes (chosen uniformly). More details about this distribution are presented in Appendix \ref{app:Design}.  


\subsection{Participants}
In total, 378 adults answered our questionnaire. Among them, 178 were recruited by advertising on private and public communication channels (i.e. LinkedIn, university networks). The rest were recruited through the Prolific crowdsourcing platform. Prolific participants had 80\% minimum approval rate and were paid \pounds3  per questionnaire, with an  average reward of \pounds9.89/h. All valid  Prolific submissions passed two attention checks.
For a real-world representation of the population, we advertised the study to targeted English language proficiency levels. As a result, 59 participants had limited working proficiency, 183 had professional proficiency, and 136 were native/bilingual. 

You can find the complete dataset with the generated adversarial sentences and the answers from the questionnaire in this link\footnote{\url{https://figshare.com/articles/dataset/ACL_2023_Human_Study_Adversarial_Text_7z/23035472}}.

%% file: results_RQ1.tex
\section{Results and Analysis}

\subsection{RQ1: Validity}
\label{sec:validity}



To 71.86\% of all adversarial examples, participants have associated the correct class label (according to the dataset ground truth). This contrasts with original examples, which human participants label correctly with 88.78\%. This difference is statistically significant (left-tailed proportion test with $Z=-12.79, p=9.92e-38$). 




Table \ref{table:rq1_attack} shows the detailed human accuracy numbers for each attack separately. Five of the nine attacks exhibit a statistical difference to original examples (the four others have over 80\% of correctly labelled adversarial examples, without significant difference with the original examples). Humans have (almost) the same accuracy as random for two of these attacks, ranging between 50 and 60\%.

\begin{table}[ht!]
\centering
\small
\begin{tabular}{@{}lcc@{}}
\toprule
Attack       & \multicolumn{1}{l}{Correctly} & Statistical difference \\ \midrule
             & \multicolumn{1}{l}{labelled}  & with original text     \\ \midrule
BAE          & 55.4                          & X                      \\
BERTAttack   & 71.1                          & X                      \\
CLARE        & 55.4                          & X                      \\
FGA          & 84.2                          & \checkmark             \\
IGA          & 87.5                          & \checkmark             \\
Kuleshov     & 86.8                          & \checkmark             \\
PSO          & 63.5                          & X                      \\
PWWS         & 74.8                          & X                      \\
TextFooler   & 85.9                          & \checkmark             \\ \midrule
All adversarial examples & 71.86                         & \checkmark             \\ 
Original & 88.78                       & -            \\ \bottomrule

\end{tabular}
\caption{Percentage of correctly labelled adversarial texts as positive or negative sentiment according to the attack method.}
\label{table:rq1_attack}
\end{table}

\textbf{Insight 1:} Five out of nine adversarial attacks generate a significant portion (>25\%) of adversarial examples that humans would interpret with the wrong label. These examples would not achieve their intended goal in human-checked NLP systems.

%% file: results_RQ2.tex
\subsection{RQ2: Naturalness}

We report below our results for the different naturalness criteria. The detailed results, globally and for each attack, are shown in Table \ref{table:naturalness_part1}.

\begin{table*}[ht!]
\small
\centering
\begin{tabular}{@{}lccccccc@{}}
\toprule
Attack                            & Suspiciousness (\%) $\downarrow$ & \multicolumn{2}{c}{Detectability(\%)$\downarrow$}                & \multicolumn{2}{c}{Grammaticality(\%) $\downarrow$} & \multicolumn{2}{c}{Meaning(1-4) $\uparrow$} \\ \midrule
\multicolumn{1}{l|}{}             & \multicolumn{1}{c|}{}            & Unknown $|\delta|$ & \multicolumn{1}{c|}{Known $|\delta|$ } & Errors exist   & \multicolumn{1}{c|}{Errors added}  & Clarity            & Preservation          \\ \cmidrule(l){3-8} 
\multicolumn{1}{l|}{BAE}          & \multicolumn{1}{c|}{50.6}        & 35.1                  & \multicolumn{1}{c|}{45.3}                & 44.2           & \multicolumn{1}{c|}{29.0}          & 2.64                  & 1.7                 \\
\multicolumn{1}{l|}{BERTAttack}   & \multicolumn{1}{c|}{63.9}        & 30.3                  & \multicolumn{1}{c|}{44.3}                & 23.7           & \multicolumn{1}{c|}{55.4}          & 2.40                  & 2.07                \\
\multicolumn{1}{l|}{CLARE}        & \multicolumn{1}{c|}{55.9}        & 45.4                  & \multicolumn{1}{c|}{39.4}                & 53.8           & \multicolumn{1}{c|}{16.4}          & 2.88                  & 1.7                 \\
\multicolumn{1}{l|}{FGA}          & \multicolumn{1}{c|}{46.5}        & 47.5                  & \multicolumn{1}{c|}{46.3}                & 44.6           & \multicolumn{1}{c|}{34.5}          & 3.06                  & 2.67                \\
\multicolumn{1}{l|}{IGA}          & \multicolumn{1}{c|}{59.1}        & 53.2                  & \multicolumn{1}{c|}{57.8}                & 36.4           & \multicolumn{1}{c|}{47.0}          & 2.70                  & 2.58                \\
\multicolumn{1}{l|}{Kuleshov}     & \multicolumn{1}{c|}{63.9}        & 57.6                  & \multicolumn{1}{c|}{65.9}                & 37.6           & \multicolumn{1}{c|}{43.9}          & 2.71                  & 2.09                \\
\multicolumn{1}{l|}{PSO}          & \multicolumn{1}{c|}{68.5}        & 46.7                  & \multicolumn{1}{c|}{54.7}                & 37.4           & \multicolumn{1}{c|}{39.1}          & 2.34                  & 1.99                \\
\multicolumn{1}{l|}{PWWS}         & \multicolumn{1}{c|}{65.5}        & 50.3                  & \multicolumn{1}{c|}{63.7}                & 34.5           & \multicolumn{1}{c|}{48.0}          & 2.26                  & 2.09                \\
\multicolumn{1}{l|}{TextFooler}   & \multicolumn{1}{c|}{61.5}        & 45.0                  & \multicolumn{1}{c|}{54.7}                & 39.1           & \multicolumn{1}{c|}{50.5}          & 2.72                  & 2.47                \\ \midrule
\multicolumn{1}{l|}{All examples} & \multicolumn{1}{c|}{60.33}       & 45.28                 & \multicolumn{1}{c|}{52.38}               & 38.9           & \multicolumn{1}{c|}{40.6}          & 2.60                  & 2.11                \\ \bottomrule
\end{tabular}
\caption{Human evaluation results about the naturalness of adversarial text.  Downwards arrows$\downarrow$  indicate lower values are preferred. Upward arrows $\uparrow$  indicate higher values are preferred. Suspicion, Detectability and Grammaticality values are percentages, while Meaning values are average of Likert scale items from 1-4. } \label{table:naturalness_part1}
\end{table*}

\subsubsection{Suspiciousness}


Humans perceive 60.33\% of adversarial examples as being computer altered. This is significantly more than the 21.43\% of the original examples that raised suspicion (right-tailed proportion test of $Z=23.63, p=9.53e^{-124}$ ). This latter percentage indicates the level of suspiciousness that attacks should target to be considered natural. A per-attack analysis (see Table \ref{table:naturalness_part1}) reveals that all attacks produce a significant number of examples perceived unnatural, from 46.55\% (FGA) to 68.5\% (PSO).

\textbf{Insight 2:}
Humans suspect that the majority of the examples (60.33\%) produced by adversarial text attacks have been altered by a computer. This demonstrates a lack of naturalness in these examples.



\subsubsection{Detectability}



When humans are not aware of the perturbation size, they can detect only 45.28\% of the altered words in examples they found to be computer altered. This percentage increases to 52.38\%, when the actual perturbation size is known (with statistical significant according to a Mann-Whitney U Test with $Z=-73.49, p=4.4e^{-8}$). These conclusions remain valid for all attacks taken individually, with a detection rate ranging from 30.3\% to 53.2\% ($\delta$ unknown) and from 39.4\% to 65.9\% ($\delta$ known). 


\textbf{Insight 3:} Humans can detect almost half (45.28\%) of the perturbed words in adversarial text. This indicates that the perturbations introduced by attacks are not imperceptible.

\subsubsection{Grammaticality}



Humans perceive grammar errors in 38.9\% of adversarial texts and claim that 40.6\% of adversarial texts contain errors not present in their original counterparts. Surprisingly, however, humans are more likely to report grammar errors in examples they perceive as original, than in those they deem computer-altered (73.0\% versus 44.6\%)(\ref{tab:grammar-suspicion}. There is thus no positive correlation between grammaticality and naturalness. 


One possible explanation is that human perception of grammar mistakes significantly differs from automated grammar checks. Indeed, the LanguageTool grammar checker \cite{naber2003rule} reports that only 17.7\% adversarial examples contain errors, which is significantly less than the 40.6\% that humans reported. This teaches us that automated grammar checks cannot substitute for human studies to assess grammaticality.

Humans report varying rates of grammar errors across different attacks. The rates are highest for CLARE (53.8\%) which is significantly more than the lowest rate (BERTAttack, 23.7\%). Human perception of the grammaticality of the different attacks changes drastically when they also see the corresponding original examples (e.g. BERTAttack has the highest error rate with 55.4\%, and CLARE has the lowest with 16.4\%), indicating again that this criterion is not relevant to explain naturalness.

Please note that the  grammar error presence and introduction are studied in two different settings (ref. section \ref{subsec:RQs} and \ref{sec:survey} ) with different sets of texts, hence can not be compared against each other. We can only comment on the results separately. 


\textbf{Insight 4:} Humans perceive grammar errors in 40\% of adversarial examples. However, there is no positive correlation between perceived grammaticality and naturalness.




\begin{table}[ht!]
\begin{tabular}{@{}llll@{}}
\toprule
                                      & Yes  & No   & Not sure \\ \midrule
\multicolumn{1}{l|}{Computer-altered} & 44.6 & 73.0 & 63.6     \\ \bottomrule
\end{tabular}
\caption{Percentage of adversarial text labelled as computer-altered according to grammar errors}
\label{tab:grammar-suspicion}
\end{table}

\subsubsection{Meaning}


Humans give an average rating of 2.60 (on a 1-4 Likert scale) to the meaning clarity of adversarial texts. This is less than original texts, which receives an average rating of 3.44 (with statistical significance based on Mann Whitney U test, with $Z=-412.10, p=1.43e^{-142}$). Furthermore, participants have mixed opinions regarding meaning preservation from original texts to adversarial texts (average rating of 2.11) on a 1-4 scale. 

To check whether lack of clarity indicates a lack of perceived naturalness, we show in Table \ref{tag:meaning-suspicion}, for each rating, the percentage of adversarial texts with this rating that humans perceived as computer altered. We observe a decreasing monotonic relation between rating and suspiciousness. This indicates that the more an adversarial text lacks clarity, the more humans are likely to consider it unnatural.

\begin{table}[ht!]
\centering
\begin{tabular}{@{}lllll@{}}
\toprule
Meaning clarity  & 1    & 2    & 3    & 4    \\ \midrule
Computer-altered & 86.8 & 75.7 & 56.7 & 25.5 \\ \bottomrule
\end{tabular}
\caption{Percentage of adversarial texts labelled as computer-altered according to clarity of meaning score}
\label{tag:meaning-suspicion}
\end{table}

All attacks have an average clarity score ranging from 2.26 (PWWS) to 3.06 (FGA), which tends to confirm the link between naturalness and meaning clarity. Meaning preservation ranges from 1.7 to 2.67. Interestingly, the attacks with a higher preservation rating (FGA, IGA, TextFooler) tends to have a higher validity score (reported in Table\ref{table:rq1_attack}), though Kuleshov is an exception.\\

\textbf{Insight 5:} Humans find adversarial text less clear than original texts, while clarity is an important factor for perceived naturalness. Moreover, attacks that preserve the original meaning tend to produce more valid examples.


%% file: results_RQ3.tex
\subsection{RQ3: How does perturbation size impact the validity and naturalness of adversarial examples?}


Pearson correlation tests have revealed that perturbation size does not affect validity and detectability, but correlates with suspiciousness, grammaticality and meaning clarity. Figure \ref{fig:perturbation} shows the graphs where a correlation was established (the others are in Appendix  \ref{app:perturbation}). Thus, adversarial examples are perceived as less natural as more word have been altered (positive correlation).  On the contrary, fewer grammatical errors are reported by humans for higher perturbations.  We performed an automated check with Language Tool, which gave the opposite results, more grammatical errors are present for larger perturbations. This again demonstrates the mismatch between human perception or knowledge of grammar errors and a predefined set of rules from automatic checkers. However, as a reminder, error presence is not the most relevant metric when evaluating adversarial text. Error introduction should be considered more important. Finally, adversarial examples with larger perturbation size have less clear meaning and preserve less original text's meaning.


\textbf{Insight 6:} The perturbation size negatively affects suspiciousness and meaning, and has no impact on validity or detectability.

\begin{figure*}
  \begin{subfigure}{0.33\linewidth}
  \includegraphics[width=\linewidth]{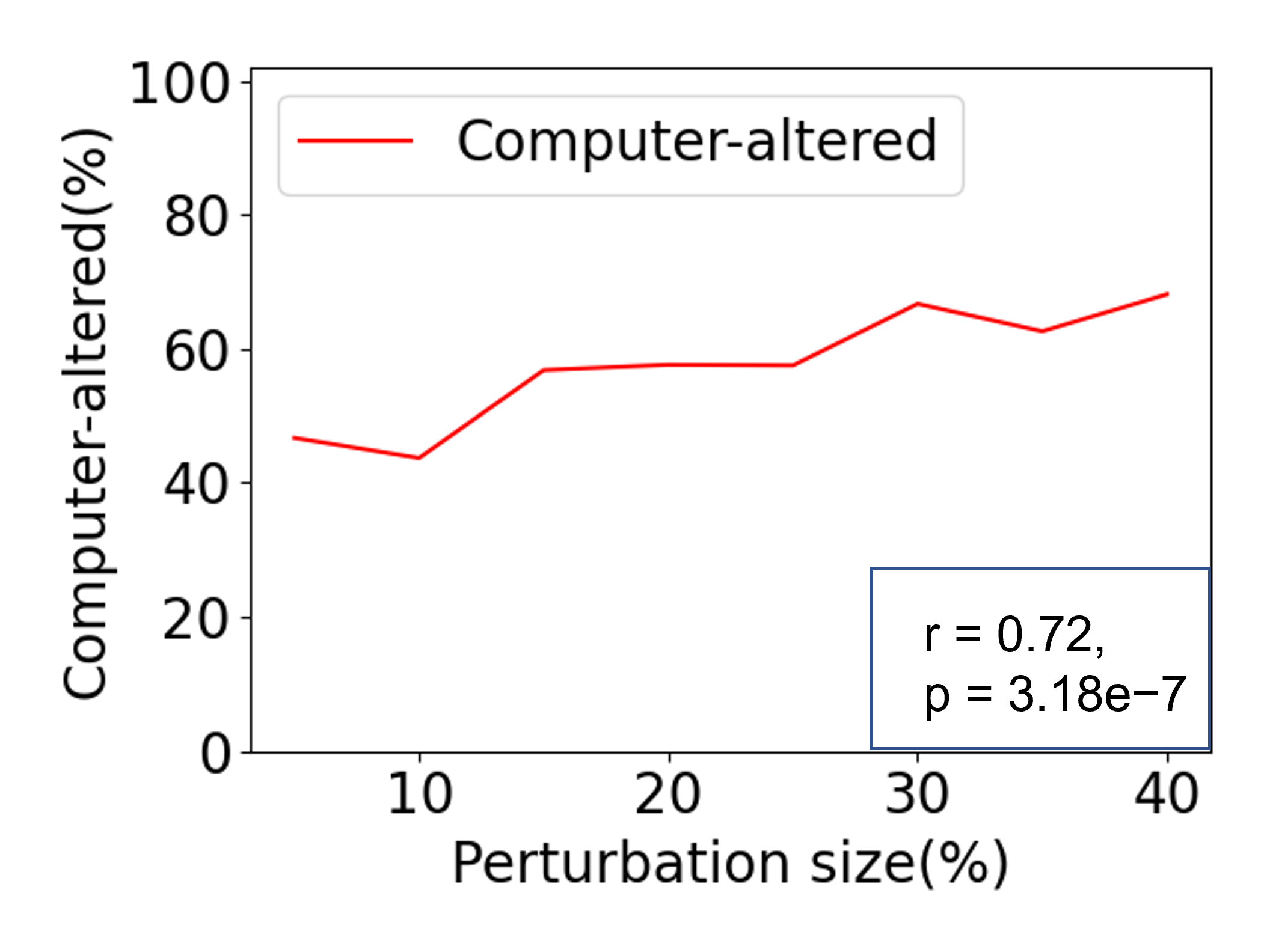}
  \subcaption{Suspiciousness}
  \end{subfigure}\hfill
  \begin{subfigure}{0.33\linewidth}
  \includegraphics[width=\linewidth]{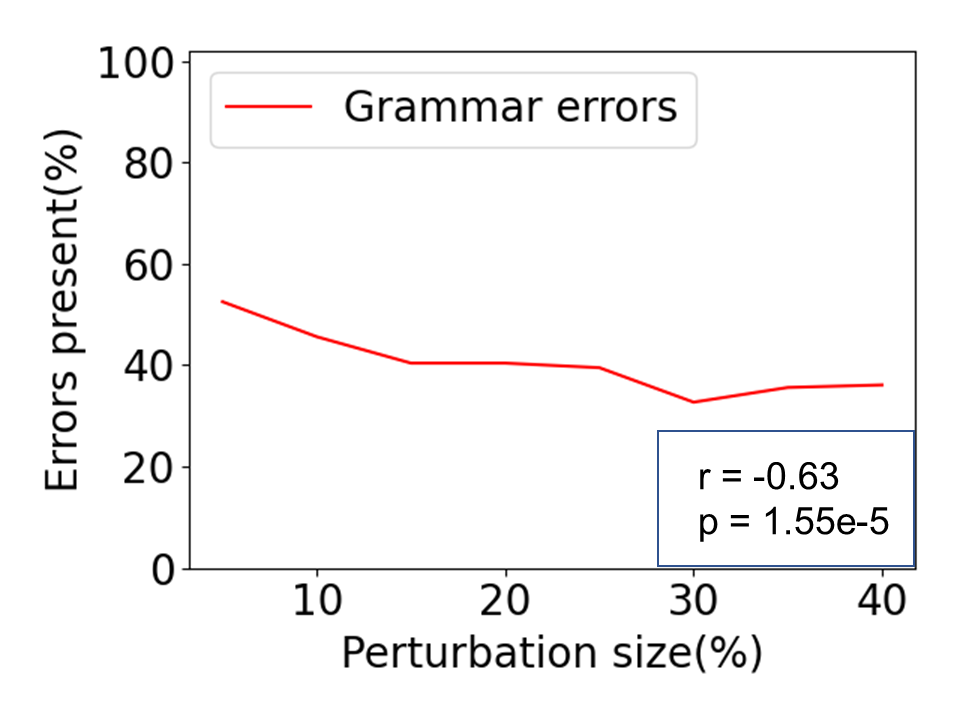}
    \subcaption{Grammaticality: Error presence}
  \end{subfigure}\hfill
    \begin{subfigure}{0.33\linewidth}
  \includegraphics[width=\linewidth]{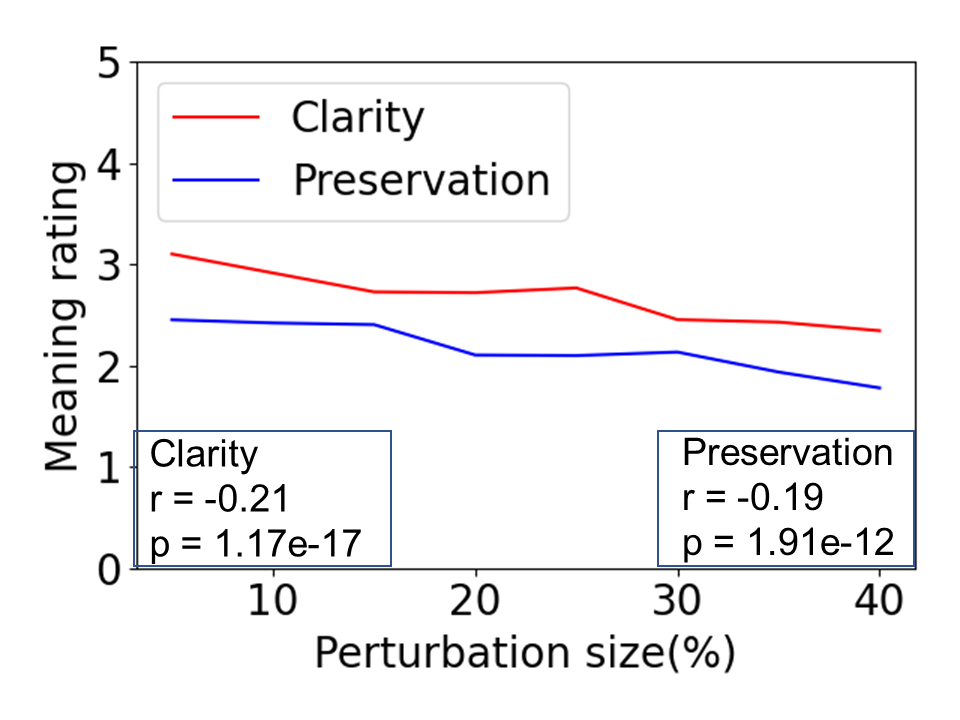}
    \subcaption[width = 0.2\linewidth]{Meaning: Clarity and Preservation}
  \end{subfigure}
  \caption{Effect of perturbation size}
  \label{fig:perturbation}
  \end{figure*}

%% file: results_misc.tex
\section{Misc. results}
We conducted an analysis to check whether human perception of naturalness and validity is related to their language proficiency. We found out that language proficiency only affects some aspects of naturalness and not validity results. People with professional proficiency are more suspicious, they achieve a higher accuracy  at detecting adversarial text compared to the other two groups(64.6\% vs 54.8\% and 57.0\%). Regarding grammaticality, people with higher proficiency level report more added errors to the original examples by adversarial attacks. Lastly, for the meaning preservation there is a statistical difference only between two proficiencies, natives give a lower score compared to limited working proficiency. For detailed results, refer to Table \ref{tab:language_app} in Appendix \label{app:language}.

%% file: Discussion.tex







%% file: Conclusion.tex
\section{Discussion and conclusion}
Our study unveils that a significant portion of adversarial examples produced by state-of-the-art text attacks would not pass human quality gates. These examples are either invalid (labelled differently from intended) or unnatural (perceived as computer altered). This means that the practical success rate of these attacks in systems interacting with humans would be lower than reported in purely model-focused evaluations.

Through our investigations, we discovered that validity is related to the meaning preservation of the original text by adversarial perturbations. As for naturalness, it appears that the detectability of (at least one) altered words, as well as meaning clarity are strong factors determining the suspiciousness of a text to have been computer-altered. The (perceived) presence of grammar errors is not a relevant criterion to determine naturalness. However, grammaticality may still make sense in contexts where exchanged texts rarely contain grammar mistakes (e.g. in professional or formal environments).

More generally, the relevant criteria to evaluate the quality of adversarial examples depend on the considered use case and threat model. Our goal, therefore, is not to qualify an existing attack as ``worse than claimed'', but rather to raise awareness that different threat scenarios may require different evaluation criteria. We, therefore, encourage researchers in adversarial attacks to precisely specify which systems and assumptions their study targets, and to justify the choice of evaluation criteria accordingly.

In particular, we corroborate previous studies that discourage the use of automated checks to replace human validation \cite{morris-etal-2020-reevaluating}. Our study has revealed that human perception of grammaticality does not match the results of grammar-checking tools. We thus argue that humans play an essential role in the evaluation of adversarial text attacks unless these attacks target specific systems that do not involve or impact humans at all.

Interestingly, none of the existing attacks dominate on all criteria. A careful observation of Tables \ref{table:rq1_attack} and \ref{table:naturalness_part1} reveals that six attacks (over nine) lie on the Pareto front (considering our evaluation criteria as objectives). This implies that different attacks fit better in different threat models.

Ultimately, we believe that our results shape relevant directions for future research on designing adversarial text. These directions include further understanding the human factors that impact the (im)perceptibility of adversarial examples, and the elaboration of new attacks optimizing these factors (in addition to model failure). The design of relevant attacks constitutes a critical step towards safer NLP models, because understanding systems' security threats paves the way for building appropriate defence mechanisms.

%% file: Limitations.tex
\section*{Limitations}

\begin{itemize}
    \item Our study focuses on word replacement attacks. While these attacks are the most common in the literature, the human perception of attacks that rely on insertion or deletion can differ from our conclusions.
    
    \item While we evaluated three datasets and over 3000 sentences, they all target the sentiment analysis classification task. \citet{MTEB} have recently released a large-scale benchmark that covers dozens of text-related tasks and datasets that can further validate our study. It would be especially interesting to consider datasets that use more formal language (i.e. journalistic).

    \item The texts we consider in this study have a maximum length of 50 words. While this allows the evaluation of a higher number of texts,  the human perception of perturbations in longer texts might differ.

    \item We considered a uniform distribution of generated adversarial texts per bin for each attack. However, their real distribution in the wild might differ from our assumed one.  
    \item All our texts and speakers revolve around the English language, while the problems that text adversarial attacks raise (such as fake news and misinformation) are global. Languages where grammar is more fluid, that allow more freedom in the positioning of the words or where subtle changes in tone significantly impact the semantics can open vulnerabilities and hence require further studies.
    
    
\end{itemize}




\section*{Ethical considerations}

This study investigates perception of humans on adversarial examples, which are modified texts that aim to change the decision of a NLP model. While these examples can be used by malicious actors, our goal is to understand the threat they bring and take informed decisions on preparing effective defences against these threats.

The texts shown to participants of this study were collected from open platforms, and it may contain inappropriate language. To mitigate this issue, we asked only participants 18+ years old to take the survey.

%% file: Appendix.tex
\appendix
\clearpage
\section{Appendices}
\label{sec:appendix}

\subsection{Distribution of texts to participants in the study}
\label{app:Design}

This study was designed to take into consideration  the level of perturbation caused to a text. As such, we use the concept  of perturbation bins, which are bins of 5\% for the perturbation size. As the maximum perturbation we study is 40\%, in total there are 8 bins. FGA and IGA attacks set a maximum perturbation size of 20\%, therefore we do not consider higher perturbations for them. 

\textbf{Dataset generation:} We split the dataset mentioned in Section \ref{dataset} in two parts: \textit{original} and \textit{adversarial}, where the original counterpart of adversarial examples in the \textit{adversarial} dataset do not intersect with the original sentences in the \textit{original} dataset. The split is done by randomly selecting first randomly 50 texts for each attack and perturbation bin combination (9x8). In the cases where the attack has generated less than 50 texts in a bin, we take all of those. In total, there were 3168 texts that were added to \textit{adversarial} dataset. To build the \textit{original} dataset, we select from the dataset in Section \ref{dataset} the original texts that are not counterparts of the texts in \textit{adversarial} dataset. Finally, the \textit{adversarial} dataset was further split in two parts by selecting randomly the  examples.

\textbf{Survey population:} We populate the survey step by step starting from Part 1. 

Part 1: Original and Adversarial text \\
We select 5 original texts randomly from \textit{original} dataset. For adversarial texts, we randomly select 5 attack - perturbation bin combinations from all possible combinations. After that, we choose 5 random texts from these 5 attack-bins from one of two sub-adversarial dataset. 

Part 2: We select for each of the 4 attacks not present in Part 1 a random perturbation bin. A random text is then picked for the given attack-bin combination from the sub-dataset of the adversarial dataset that was not picked in Part 1. 

Part3: The same adversarial texts as in Part 2, joined with their original counterparts.

\begin{figure*}[ht!]
  \includegraphics[width=\linewidth]{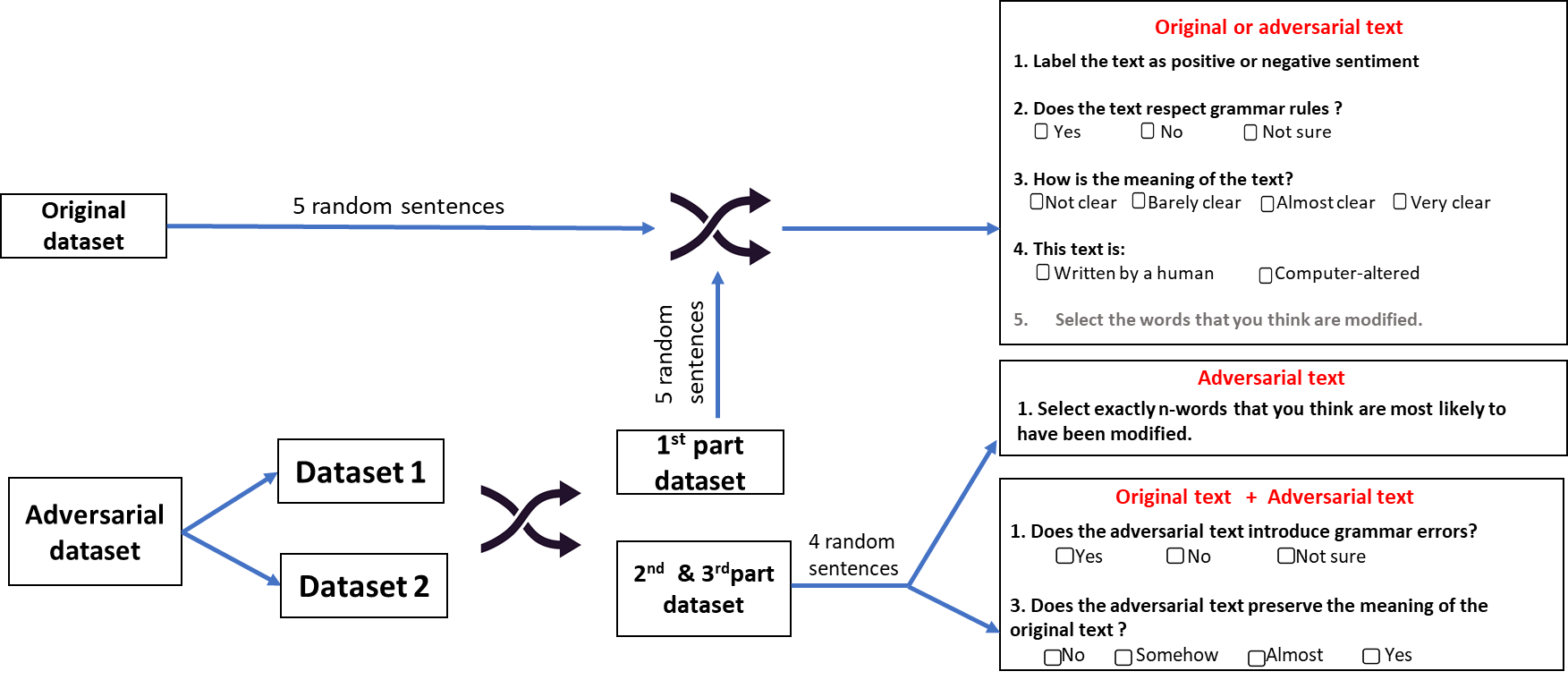}
  \caption{Distribution procedure of texts to participants of the questionnaire.}
  \label{fig:distribution}
  \end{figure*}

The full distribution of the texts to participants  is illustrated in Figure \ref{fig:distribution}. The distribution of answers per attack and bin is given in Table \ref{tab:answer_dist_bin1}  and \ref{tab:answer_dist_bin2} .

\begin{table}[ht!]
\small
\centering
\begin{tabular}{@{}lllllllll@{}}
\toprule
Attack \textbackslash  \quad Bin & 5  & 10 & 15 & 20 & 25 & 30 & 35 & 40 \\ \hline
BAE          & 16 & 11 & 19 & 31 & 38 & 29 & 27 & 62 \\
BERT         & 11 & 8  & 17 & 15 & 22 & 16 & 29 & 76 \\
CLARE        & 9  & 13 & 21 & 50 & 30 & 11 & 20 & 41 \\
FGAJia       & 14 & 18 & 22 & 47 & 0  & 0  & 0  & 0  \\
IGAWang      & 13 & 17 & 22 & 36 & 0  & 0  & 0  & 0  \\
Kuleshov     & 16 & 7  & 10 & 14 & 29 & 27 & 42 & 60 \\
PSO          & 15 & 10 & 11 & 17 & 23 & 28 & 42 & 76 \\
PWWS         & 13 & 12 & 17 & 19 & 30 & 30 & 33 & 72 \\
TextFooler   & 13 & 7  & 7  & 16 & 28 & 27 & 29 & 65 \\ \hline
\end{tabular}
\caption{Number of evaluations according to bins for Part 1 of the questionnaire}
\label{tab:answer_dist_bin1}
\end{table}

\begin{table}[ht!]
\small
\centering
\begin{tabular}{@{}lllllllll@{}}
\toprule
Attack \textbackslash  \quad Bin      & 5  & 10 & 15 & 20 & 25 & 30 & 35 & 40 \\ \midrule
BAE      & 7  & 8  & 11 & 17 & 33 & 14 & 26 & 29 \\
BERTAttack  & 12 & 13 & 13 & 21 & 32 & 24 & 29 & 40 \\
CLARE        & 15 & 18 & 24 & 55 & 31 & 11 & 16 & 13 \\
FGA      & 12 & 8  & 12 & 26 & 0  & 0  & 0  & 0  \\
IGA       & 6  & 12 & 22 & 26 & 0  & 0  & 0  & 0  \\
Kuleshov      & 12 & 6  & 14 & 29 & 27 & 24 & 32 & 29 \\
PSO       & 8  & 5  & 14 & 18 & 19 & 31 & 25 & 36 \\
PWWSRen      & 8  & 7  & 12 & 21 & 22 & 23 & 33 & 26 \\
TextFooler & 11 & 16 & 17 & 20 & 21 & 32 & 39 & 30 \\ \bottomrule
\end{tabular}
\caption{Number of evaluations according to bins for Part 2 and 3 of the questionnaire}
\label{tab:answer_dist_bin2}
\end{table}

\subsection{Effect of perturbation size}
\label{app:perturbation}
\begin{figure*}[ht!]
  \begin{subfigure}{0.33\linewidth}
  \includegraphics[trim={0, 0, 0, 0}, width=\linewidth]{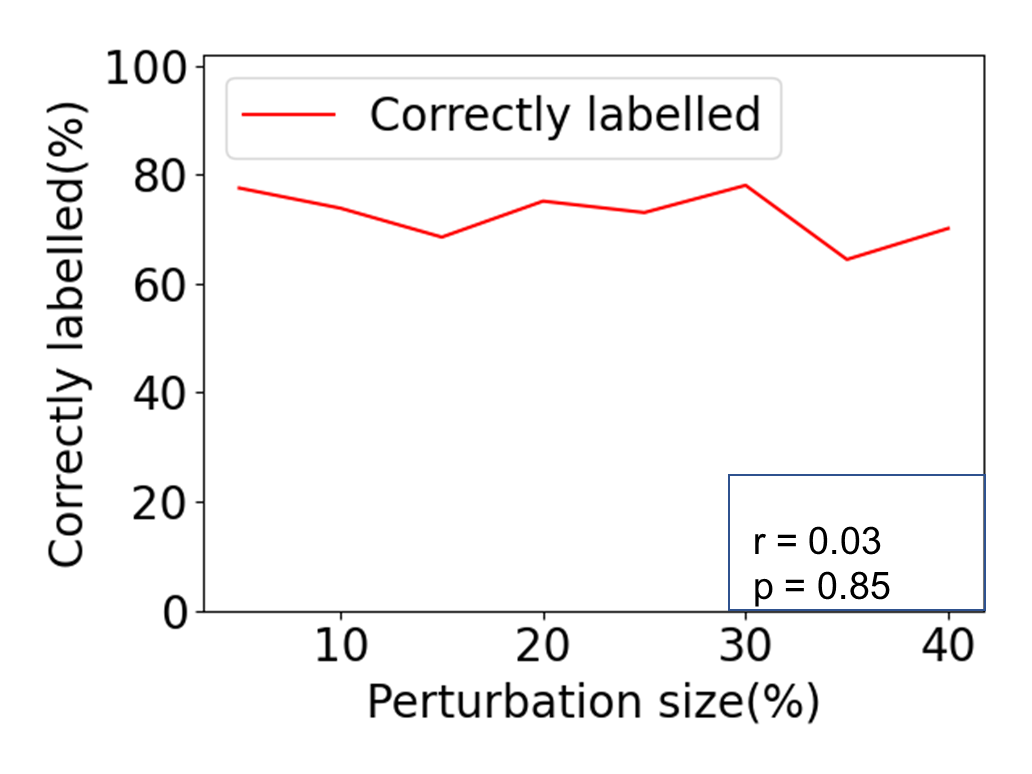}
  \subcaption{Validity}
  \end{subfigure}\hfill 
  \begin{subfigure}{0.33\linewidth}
  \includegraphics[trim={0, 0, 0, 0}, width=\linewidth]{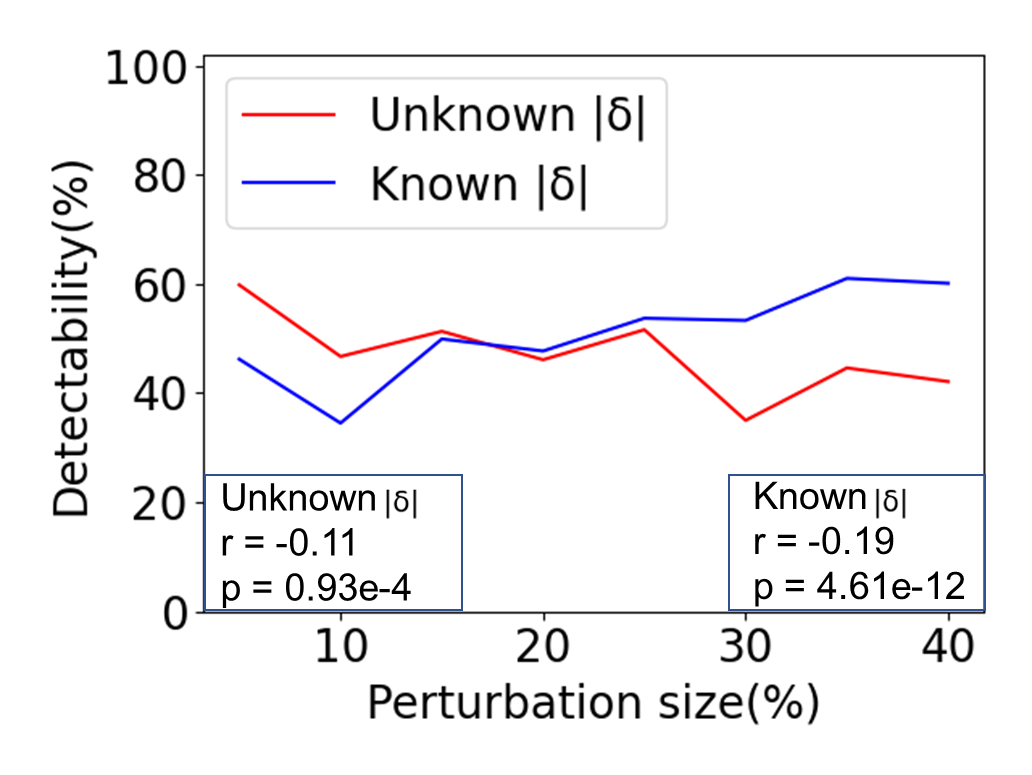}
    \caption{Detectability}
  \end{subfigure}\hfill 
    \begin{subfigure}{0.33\linewidth}
  \includegraphics[trim={0, 0, 0, 0}, width=\linewidth]{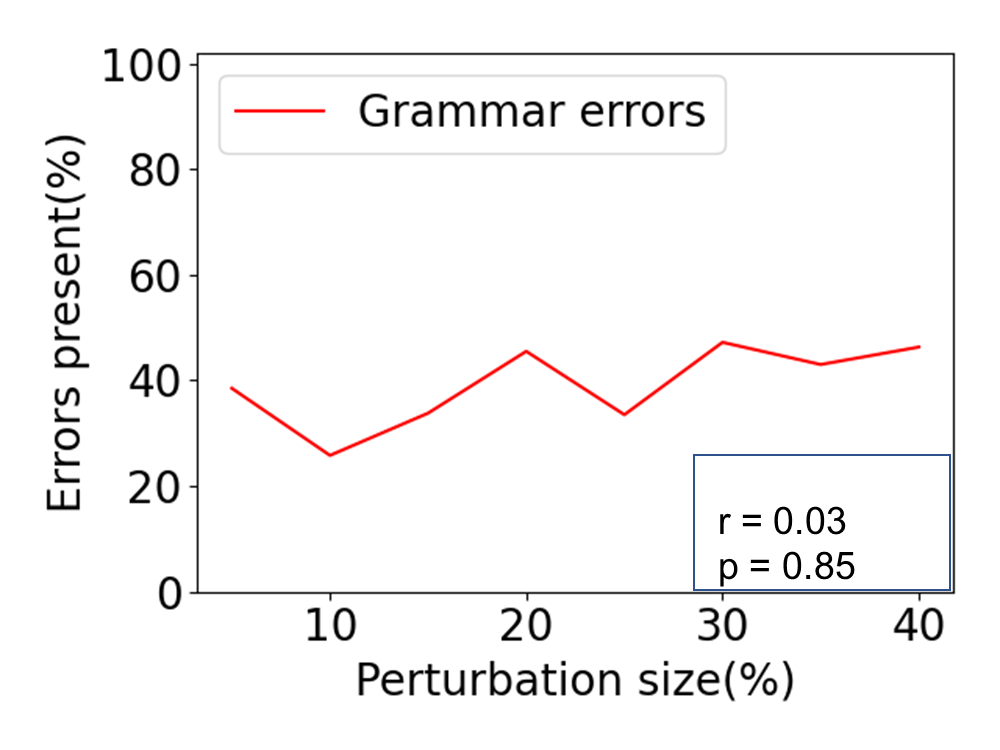}
    \caption{Grammaticality}
  \end{subfigure}
  \caption{Effect of perturbation size}
  \label{fig:perturbation_app}
  \end{figure*}

Conducting Pearson correlation tests, we found that perturbation size does not affect validity, detectability (unknown and known perturbation size) and grammar errors introduced by perturbations. Figure \ref{fig:perturbation_app} shows visually the relationship as well as test statistics.

\subsection{Language proficiency effect}
\label{app:language}
Table \ref{tab:language_app} shows the effect of language proficiency in the evaluated metrics for naturality and validity.

\begin{table*}[ht!]
\scriptsize
\centering
\begin{tabular}{@{}lcccccccc@{}}
\toprule
Proficiency                       & \multicolumn{1}{l}{Validity(\%)$\downarrow$} & Suspicion (\%) $\downarrow$ & \multicolumn{2}{c}{Detectability(\%)$\downarrow$}                & \multicolumn{2}{c}{Grammaticality(\%) $\downarrow$} & \multicolumn{2}{c}{Meaning(1-4) $\uparrow$} \\ \midrule
\multicolumn{1}{l|}{}             & \multicolumn{1}{l|}{}                       & \multicolumn{1}{c|}{}            & Unknown |$\delta$| & \multicolumn{1}{c|}{Known |$\delta$|} & Errors exist   & \multicolumn{1}{c|}{Errors added}  & Clarity             & Preservation            \\ \cmidrule(l){4-9} 
\multicolumn{1}{l|}{Limited}      & \multicolumn{1}{c|}{72.4}                   & \multicolumn{1}{c|}{54.8}        & 42.7                  & \multicolumn{1}{c|}{48.6}                & 38.8           & \multicolumn{1}{c|}{31.7}          & 2.65               & 2.26                   \\
\multicolumn{1}{l|}{Professional} & \multicolumn{1}{c|}{72.6}                   & \multicolumn{1}{c|}{64.6}        & 43.9                  & \multicolumn{1}{c|}{52.5}                & 38.3           & \multicolumn{1}{c|}{40.1}          & 2.63               & 2                      \\
\multicolumn{1}{l|}{Native}       & \multicolumn{1}{c|}{70.7}                   & \multicolumn{1}{c|}{57.0}        & 48.4                  & \multicolumn{1}{c|}{53.8}                & 39.7           & \multicolumn{1}{c|}{45.0}          & 2.55               & 2.15                   \\ \midrule
\multicolumn{1}{l|}{All examples} & \multicolumn{1}{c|}{71.36}                  & \multicolumn{1}{c|}{60.33}       & 45.28                 & \multicolumn{1}{c|}{52.38}               & 38.9           & \multicolumn{1}{c|}{40.6}          & 2.60               & 2.11                   \\ \bottomrule
\end{tabular}
\caption{Effect of language proficiency on the perception of validity and naturalness for human participants.}
\label{tab:language_app}
\end{table*}

%% file: acl2023.bbl
\begin{thebibliography}{32}
\expandafter\ifx\csname natexlab\endcsname\relax\def\natexlab#1{#1}\fi

\bibitem[{Ali et~al.(2021)Ali, Khan, AlGhadhban, Alazmi, Alzamil, Al-Utaibi,
  and Qadir}]{ali2021all}
Hassan Ali, Muhammad~Suleman Khan, Amer AlGhadhban, Meshari Alazmi, Ahmad
  Alzamil, Khaled Al-Utaibi, and Junaid Qadir. 2021.
\newblock All your fake detector are belong to us: Evaluating adversarial
  robustness of fake-news detectors under black-box settings.
\newblock \emph{IEEE Access}, 9:81678--81692.

\bibitem[{Alzantot et~al.(2018)Alzantot, Sharma, Elgohary, Ho, Srivastava, and
  Chang}]{alzantot-etal-2018-generating}
Moustafa Alzantot, Yash Sharma, Ahmed Elgohary, Bo-Jhang Ho, Mani Srivastava,
  and Kai-Wei Chang. 2018.
\newblock \href {https://doi.org/10.18653/v1/D18-1316} {Generating natural
  language adversarial examples}.
\newblock In \emph{Proceedings of the 2018 Conference on Empirical Methods in
  Natural Language Processing}, pages 2890--2896, Brussels, Belgium.
  Association for Computational Linguistics.

\bibitem[{Chen et~al.(2022)Chen, Gao, Cui, Qi, Huang, Liu, and
  Sun}]{chen-etal-2022-adversarial}
Yangyi Chen, Hongcheng Gao, Ganqu Cui, Fanchao Qi, Longtao Huang, Zhiyuan Liu,
  and Maosong Sun. 2022.
\newblock \href {https://aclanthology.org/2022.emnlp-main.771} {Why should
  adversarial perturbations be imperceptible? rethink the research paradigm in
  adversarial {NLP}}.
\newblock In \emph{Proceedings of the 2022 Conference on Empirical Methods in
  Natural Language Processing}, pages 11222--11237, Abu Dhabi, United Arab
  Emirates. Association for Computational Linguistics.

\bibitem[{Dyrmishi et~al.(2022)Dyrmishi, Ghamizi, Simonetto, Traon, and
  Cordy}]{dyrmishi2022empirical}
Salijona Dyrmishi, Salah Ghamizi, Thibault Simonetto, Yves~Le Traon, and Maxime
  Cordy. 2022.
\newblock On the empirical effectiveness of unrealistic adversarial hardening
  against realistic adversarial attacks.
\newblock \emph{arXiv preprint arXiv:2202.03277}.

\bibitem[{Ebrahimi et~al.(2018)Ebrahimi, Rao, Lowd, and
  Dou}]{ebrahimi-etal-2018-hotflip}
Javid Ebrahimi, Anyi Rao, Daniel Lowd, and Dejing Dou. 2018.
\newblock \href {https://doi.org/10.18653/v1/P18-2006} {{H}ot{F}lip: White-box
  adversarial examples for text classification}.
\newblock In \emph{Proceedings of the 56th Annual Meeting of the Association
  for Computational Linguistics (Volume 2: Short Papers)}, pages 31--36,
  Melbourne, Australia. Association for Computational Linguistics.

\bibitem[{Feng et~al.(2018)Feng, Wallace, Grissom~II, Iyyer, Rodriguez, and
  Boyd-Graber}]{feng-etal-2018-pathologies}
Shi Feng, Eric Wallace, Alvin Grissom~II, Mohit Iyyer, Pedro Rodriguez, and
  Jordan Boyd-Graber. 2018.
\newblock \href {https://doi.org/10.18653/v1/D18-1407} {Pathologies of neural
  models make interpretations difficult}.
\newblock In \emph{Proceedings of the 2018 Conference on Empirical Methods in
  Natural Language Processing}, pages 3719--3728, Brussels, Belgium.
  Association for Computational Linguistics.

\bibitem[{Gao et~al.(2018)Gao, Lanchantin, Soffa, and Qi}]{gao2018black}
Ji~Gao, Jack Lanchantin, Mary~Lou Soffa, and Yanjun Qi. 2018.
\newblock Black-box generation of adversarial text sequences to evade deep
  learning classifiers.
\newblock In \emph{2018 IEEE Security and Privacy Workshops (SPW)}, pages
  50--56. IEEE.

\bibitem[{Garg and Ramakrishnan(2020)}]{garg-ramakrishnan-2020-bae}
Siddhant Garg and Goutham Ramakrishnan. 2020.
\newblock \href {https://doi.org/10.18653/v1/2020.emnlp-main.498} {{BAE}:
  {BERT}-based adversarial examples for text classification}.
\newblock In \emph{Proceedings of the 2020 Conference on Empirical Methods in
  Natural Language Processing (EMNLP)}, pages 6174--6181, Online. Association
  for Computational Linguistics.

\bibitem[{Iyyer et~al.(2018)Iyyer, Wieting, Gimpel, and
  Zettlemoyer}]{iyyer2018adversarial}
Mohit Iyyer, John Wieting, Kevin Gimpel, and Luke Zettlemoyer. 2018.
\newblock Adversarial example generation with syntactically controlled
  paraphrase networks.
\newblock \emph{arXiv preprint arXiv:1804.06059}.

\bibitem[{Jia et~al.(2019)Jia, Raghunathan, G{\"o}ksel, and
  Liang}]{jia-etal-2019-certified}
Robin Jia, Aditi Raghunathan, Kerem G{\"o}ksel, and Percy Liang. 2019.
\newblock \href {https://doi.org/10.18653/v1/D19-1423} {Certified robustness to
  adversarial word substitutions}.
\newblock In \emph{Proceedings of the 2019 Conference on Empirical Methods in
  Natural Language Processing and the 9th International Joint Conference on
  Natural Language Processing (EMNLP-IJCNLP)}, pages 4129--4142, Hong Kong,
  China. Association for Computational Linguistics.

\bibitem[{Jin et~al.(2020)Jin, Jin, Zhou, and Szolovits}]{jin2020bert}
Di~Jin, Zhijing Jin, Joey~Tianyi Zhou, and Peter Szolovits. 2020.
\newblock Is bert really robust? a strong baseline for natural language attack
  on text classification and entailment.
\newblock In \emph{Proceedings of the AAAI conference on artificial
  intelligence}, volume~34, pages 8018--8025.

\bibitem[{Kuchipudi et~al.(2020)Kuchipudi, Nannapaneni, and
  Liao}]{kuchipudi2020adversarial}
Bhargav Kuchipudi, Ravi~Teja Nannapaneni, and Qi~Liao. 2020.
\newblock Adversarial machine learning for spam filters.
\newblock In \emph{Proceedings of the 15th International Conference on
  Availability, Reliability and Security}, pages 1--6.

\bibitem[{Kuleshov et~al.(2018)Kuleshov, Thakoor, Lau, and
  Ermon}]{kuleshov2018adversarial}
Volodymyr Kuleshov, Shantanu Thakoor, Tingfung Lau, and Stefano Ermon. 2018.
\newblock Adversarial examples for natural language classification problems.

\bibitem[{Li et~al.(2021)Li, Zhang, Peng, Chen, Brockett, Sun, and
  Dolan}]{li-etal-2021-contextualized}
Dianqi Li, Yizhe Zhang, Hao Peng, Liqun Chen, Chris Brockett, Ming-Ting Sun,
  and Bill Dolan. 2021.
\newblock \href {https://doi.org/10.18653/v1/2021.naacl-main.400}
  {Contextualized perturbation for textual adversarial attack}.
\newblock In \emph{Proceedings of the 2021 Conference of the North American
  Chapter of the Association for Computational Linguistics: Human Language
  Technologies}, pages 5053--5069, Online. Association for Computational
  Linguistics.

\bibitem[{Li et~al.(2019)Li, Ji, Du, Li, and Wang}]{Li2018TextBuggerGA}
Jinfeng Li, Shouling Ji, Tianyu Du, Bo~Li, and Ting Wang. 2019.
\newblock Textbugger: Generating adversarial text against real-world
  applications.
\newblock In \emph{NETWORK AND DISTRIBUTED SYSTEM SECURITY SYMPOSIUM}, pages
  391--406.

\bibitem[{Li et~al.(2020)Li, Ma, Guo, Xue, and Qiu}]{li-etal-2020-bert-attack}
Linyang Li, Ruotian Ma, Qipeng Guo, Xiangyang Xue, and Xipeng Qiu. 2020.
\newblock \href {https://doi.org/10.18653/v1/2020.emnlp-main.500}
  {{BERT}-{ATTACK}: Adversarial attack against {BERT} using {BERT}}.
\newblock In \emph{Proceedings of the 2020 Conference on Empirical Methods in
  Natural Language Processing (EMNLP)}, pages 6193--6202, Online. Association
  for Computational Linguistics.

\bibitem[{Maas et~al.(2011)Maas, Daly, Pham, Huang, Ng, and
  Potts}]{maas-EtAl:2011:ACL-HLT2011}
Andrew~L. Maas, Raymond~E. Daly, Peter~T. Pham, Dan Huang, Andrew~Y. Ng, and
  Christopher Potts. 2011.
\newblock \href {http://www.aclweb.org/anthology/P11-1015} {Learning word
  vectors for sentiment analysis}.
\newblock In \emph{Proceedings of the 49th Annual Meeting of the Association
  for Computational Linguistics: Human Language Technologies}, pages 142--150,
  Portland, Oregon, USA. Association for Computational Linguistics.

\bibitem[{Michel et~al.(2019)Michel, Li, Neubig, and
  Pino}]{michel-etal-2019-evaluation}
Paul Michel, Xian Li, Graham Neubig, and Juan Pino. 2019.
\newblock \href {https://doi.org/10.18653/v1/N19-1314} {On evaluation of
  adversarial perturbations for sequence-to-sequence models}.
\newblock In \emph{Proceedings of the 2019 Conference of the North {A}merican
  Chapter of the Association for Computational Linguistics: Human Language
  Technologies, Volume 1 (Long and Short Papers)}, pages 3103--3114,
  Minneapolis, Minnesota. Association for Computational Linguistics.

\bibitem[{Morris et~al.(2020{\natexlab{a}})Morris, Lifland, Lanchantin, Ji, and
  Qi}]{morris-etal-2020-reevaluating}
John Morris, Eli Lifland, Jack Lanchantin, Yangfeng Ji, and Yanjun Qi.
  2020{\natexlab{a}}.
\newblock \href {https://doi.org/10.18653/v1/2020.findings-emnlp.341}
  {Reevaluating adversarial examples in natural language}.
\newblock In \emph{Findings of the Association for Computational Linguistics:
  EMNLP 2020}, pages 3829--3839, Online. Association for Computational
  Linguistics.

\bibitem[{Morris et~al.(2020{\natexlab{b}})Morris, Lifland, Yoo, Grigsby, Jin,
  and Qi}]{morris2020textattack}
John Morris, Eli Lifland, Jin~Yong Yoo, Jake Grigsby, Di~Jin, and Yanjun Qi.
  2020{\natexlab{b}}.
\newblock Textattack: A framework for adversarial attacks, data augmentation,
  and adversarial training in nlp.
\newblock In \emph{Proceedings of the 2020 Conference on Empirical Methods in
  Natural Language Processing: System Demonstrations}, pages 119--126.

\bibitem[{Muennighoff et~al.(2022)Muennighoff, Tazi, Magne, and Reimers}]{MTEB}
Niklas Muennighoff, Nouamane Tazi, Loïc Magne, and Nils Reimers. 2022.
\newblock \href {https://doi.org/10.48550/ARXIV.2210.07316} {Mteb: Massive text
  embedding benchmark}.

\bibitem[{Naber et~al.(2003)}]{naber2003rule}
Daniel Naber et~al. 2003.
\newblock A rule-based style and grammar checker.

\bibitem[{Pang and Lee(2005)}]{Pang+Lee:05a}
Bo~Pang and Lillian Lee. 2005.
\newblock Seeing stars: Exploiting class relationships for sentiment
  categorization with respect to rating scales.
\newblock In \emph{Proceedings of the ACL}.

\bibitem[{Pruthi et~al.(2019)Pruthi, Dhingra, and
  Lipton}]{pruthi-etal-2019-combating}
Danish Pruthi, Bhuwan Dhingra, and Zachary~C. Lipton. 2019.
\newblock \href {https://doi.org/10.18653/v1/P19-1561} {Combating adversarial
  misspellings with robust word recognition}.
\newblock In \emph{Proceedings of the 57th Annual Meeting of the Association
  for Computational Linguistics}, pages 5582--5591, Florence, Italy.
  Association for Computational Linguistics.

\bibitem[{Qi et~al.(2021)Qi, Chen, Zhang, Li, Liu, and Sun}]{qi-etal-2021-mind}
Fanchao Qi, Yangyi Chen, Xurui Zhang, Mukai Li, Zhiyuan Liu, and Maosong Sun.
  2021.
\newblock \href {https://doi.org/10.18653/v1/2021.emnlp-main.374} {Mind the
  style of text! adversarial and backdoor attacks based on text style
  transfer}.
\newblock In \emph{Proceedings of the 2021 Conference on Empirical Methods in
  Natural Language Processing}, pages 4569--4580, Online and Punta Cana,
  Dominican Republic. Association for Computational Linguistics.

\bibitem[{Ren et~al.(2019)Ren, Deng, He, and Che}]{ren2019generating}
Shuhuai Ren, Yihe Deng, Kun He, and Wanxiang Che. 2019.
\newblock Generating natural language adversarial examples through probability
  weighted word saliency.
\newblock In \emph{Proceedings of the 57th annual meeting of the association
  for computational linguistics}, pages 1085--1097.

\bibitem[{Sakaguchi et~al.(2017)Sakaguchi, Post, and
  Van~Durme}]{sakaguchi-etal-2017-grammatical}
Keisuke Sakaguchi, Matt Post, and Benjamin Van~Durme. 2017.
\newblock \href {https://aclanthology.org/I17-2062} {Grammatical error
  correction with neural reinforcement learning}.
\newblock In \emph{Proceedings of the Eighth International Joint Conference on
  Natural Language Processing (Volume 2: Short Papers)}, pages 366--372,
  Taipei, Taiwan. Asian Federation of Natural Language Processing.

\bibitem[{Simonetto et~al.(2021)Simonetto, Dyrmishi, Ghamizi, Cordy, and
  Traon}]{constrained_adversarials}
Thibault Simonetto, Salijona Dyrmishi, Salah Ghamizi, Maxime Cordy, and Yves~Le
  Traon. 2021.
\newblock \href {https://doi.org/10.48550/ARXIV.2112.01156} {A unified
  framework for adversarial attack and defense in constrained feature space}.

\bibitem[{Wang et~al.(2019)Wang, Jin, and He}]{wang2019natural}
Xiaosen Wang, Hao Jin, and Kun He. 2019.
\newblock Natural language adversarial attacks and defenses in word level.
\newblock \emph{arXiv preprint arXiv:1909.06723}.

\bibitem[{Zang et~al.(2020)Zang, Qi, Yang, Liu, Zhang, Liu, and
  Sun}]{zang2020word}
Yuan Zang, Fanchao Qi, Chenghao Yang, Zhiyuan Liu, Meng Zhang, Qun Liu, and
  Maosong Sun. 2020.
\newblock Word-level textual adversarial attacking as combinatorial
  optimization.
\newblock In \emph{Proceedings of ACL}.

\bibitem[{Zeng et~al.(2021)Zeng, Qi, Zhou, Zhang, Hou, Zang, Liu, and
  Sun}]{zeng2020openattack}
Guoyang Zeng, Fanchao Qi, Qianrui Zhou, Tingji Zhang, Bairu Hou, Yuan Zang,
  Zhiyuan Liu, and Maosong Sun. 2021.
\newblock \href {https://doi.org/10.18653/v1/2021.acl-demo.43} {{Openattack: An
  open-source textual adversarial attack toolkit}}.
\newblock In \emph{Proceedings of the 59th Annual Meeting of the Association
  for Computational Linguistics and the 11th International Joint Conference on
  Natural Language Processing: System Demonstrations}, pages 363--371.

\bibitem[{Zhang et~al.(2015)Zhang, Zhao, and LeCun}]{zhang2015character}
Xiang Zhang, Junbo Zhao, and Yann LeCun. 2015.
\newblock Character-level convolutional networks for text classification.
\newblock \emph{Advances in neural information processing systems}, 28.

\end{thebibliography}
